\documentclass{article}
\usepackage{spconf,amsmath,graphicx,hyperref}

\title{Resolution Allocation with Learnable Resizing and Spectral Filter Parameterization}

\name{Wei Wu, Eric C. Larson}
\address{Department of Computer Science, Southern Methodist University, Dallas, TX, USA}

\begin{document}

\maketitle

\begin{abstract}

Learnable downsampling and spectral-domain filter parameterizations have each made convolutional networks more flexible, but their interaction has not been systematically examined. We bridge this gap by examining convolutional kernels learned in the spatial and spectral domains, together with learned downsampling ratios under a cropped resizing implementation. We find that spectral and spatial kernels respond differently to learned resolution allocation, with larger gains for the spectral kernel. We observe this consistently on three datasets (EuroSAT, CIFAR-10 and STL-10) and at two model sizes (differences in accuracy gain of +0.013 to +0.028 on held-out test sets). Part of this difference reflects the greater gain of the spectral kernel from retaining more spatial area, but the difference remains positive when total retained area is matched. This pattern is observed across datasets and model sizes, although its magnitude and dependence on the complexity budget vary with model size and the resizing implementation.

\end{abstract}


\section{Introduction}
\label{sec:intro}

In convolutional networks, spatial resolution is usually fixed by architectural choices such as strides, pooling windows and resizing ratios. Recent work makes it learnable: DiffStride~\cite{diffstride} learns strides through differentiable cropping in the Fourier domain, Resolution Layers~\cite{harper2026} extend this to both downsampling and upsampling without tying it to a particular transform, and DNArch~\cite{dnarch} learns downsampling jointly with continuous kernels generated by an MLP. These methods are evaluated mainly by predictive performance, not by how learned resolution interacts with the way the filters are parameterized.

A separate line of work learns convolutional filters in the spectral domain~\cite{rippel2015spectral,pratt2017fcnn,wood2021parametric,harper2025scaling}, which gives control over the spatial extent of the filtering operation. DNArch~\cite{dnarch} conjectures that generated kernels may be easier to construct at lower resolutions, suggesting an interaction between kernel parameterization and resolution allocation, but this has not been tested.

To examine how these methods interact, we combine learnable resizing~\cite{harper2026} with either $3 \times 3$ spatial kernels or spectral kernels generated by a coordinate-based MLP~\cite{harper2025scaling}, while controlling the cutoff-frequency budget and, in our main comparison, approximately matching retained feature-map area between kernel types. Our results establish three observations. First, spectral and spatial kernels respond differently to learned resolution allocation: spectral kernels obtain significantly larger gains than spatial kernels. This difference is observed across three datasets and two model sizes. The difference remains when the comparison is made against a fixed schedule with the same total retained area, although the effect is smaller. Second, the magnitude of this difference decreases as model size increases, under both the soft and the cropped implementations. Third, the magnitude of the effect depends on how resizing is implemented: a full-resolution soft-masking approach produces a larger benefit (but requires substantially more computation), whereas the cropped implementation produces a smaller (but measurable) effect. These results suggest that spatial resolution cannot be treated independently of filter parameterization.

\begin{figure}[t]
\centering
\includegraphics[width=\linewidth]{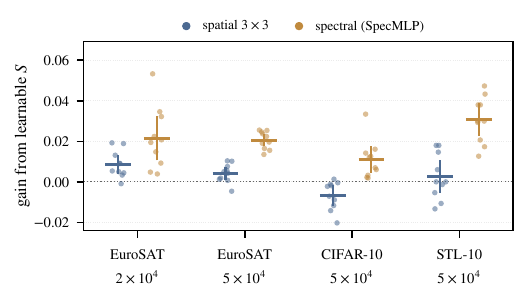}
\caption{Gain from making $S$ learnable, by filter type, under the cropped implementation (learnable resizing with feature maps cropped where the mask falls below 0.01). Held-out test accuracy relative to the fixed schedule at $p^\star = 0.12$; points are individual seeds, horizontal bars are means and vertical bars are 95\% confidence intervals ($n = 10$). The spectral filters gain more in all four settings.}
\label{fig:interaction}
\end{figure}

\section{Related work}
\label{sec:related}

This work builds on learnable downsampling and learnable kernel sizes. DiffStride~\cite{diffstride} learns fractional strides by cropping the Fourier spectrum under a smooth adaptive-span mask, while Resolution Layers~\cite{harper2026} extend learnable resizing to downsampling, upsampling, and other transforms. Shape Adaptor~\cite{liu2020shape} and DynOPool~\cite{jang2022dynopool} learn task-dependent feature-map scale factors, the latter under a complexity constraint. These methods optimize resolution or scale but do not study its interaction with filter parameterization.

DNArch~\cite{dnarch} learns frequency-domain downsampling jointly with MLP-generated kernels and conjectures that such kernels are easier to construct at lower resolutions, but does not test whether filter parameterization changes the benefit of learned resolution. We instantiate truncation as in Resolution Layers, soft masking as in DNArch, and ramp cropping as in DiffStride in one backbone (Sec.~3.2). Downsampling can introduce aliasing~\cite{ribeiro2021aliasing}, while differentiable discrete choices follow the broader DARTS framework~\cite{darts}.

\section{Method}
\label{sec:method}

\subsection{Complexity Budget}

Let the network contain $L$ resize layers, with $s_l$ denoting the DCT cutoff at layer $l$. Since discarded resolution cannot be recovered, the effective resolution is $\tilde{s}_l=\min_{j\leq l}s_j$. We define
\[
C=\sum_{l=1}^{L}\tilde{s}_l^2,
\]
since $\tilde{s}_l^2$ is proportional to the retained feature-map area, and normalize by $C_{\mathrm{full}}$, obtained when $s_l=H$ for all $l$. Thus $C/C_{\mathrm{full}}$ is a proxy for retained spatial area rather than total end-to-end FLOPs; actual feature-map area also depends on the resizing implementation (Sec.~3.2).

The training objective is the categorical cross-entropy loss $\mathcal{L}_\mathrm{cls}$ augmented with a quadratic penalty that pulls the realized complexity toward a target fraction $p^\star \in (0,1]$,
\begin{equation}
    \mathcal{L} = \mathcal{L}_\mathrm{cls}
    + \lambda \left( \frac{C}{C_\mathrm{full}} - p^\star \right)^2 
    \label{eq:loss}
\end{equation}
We set $\lambda = 100$ so that the realized complexity stays close to the target throughout training, ensuring that different configurations are compared under the same budget. Our main comparisons use $p^\star = 0.12$; Sec.~4.4 sweeps seven budgets between $0.08$ and $0.50$.

\subsection{Learnable Resizing}

Each of the $L = 4$ resizing layers, one before each stage, is parameterized by a scalar $s_l$, and we write $S = (s_1, \dots, s_4)$. A layer applies a 2D discrete cosine transform (DCT), multiplies the coefficients by a mask, and applies an inverse DCT. The 1D mask is
\begin{equation} 
m_k = \sigma\!\left(\frac{s - k}{\tau}\right),
\end{equation}
where $k$ is the frequency index and $\tau = 2$; the 2D mask is $\mathbf{M} = \mathbf{m} \otimes \mathbf{m}$. Because $s$ appears inside the sigmoid, it receives gradients and the effective resolution becomes learnable. We compare three implementations.

\textbf{Fixed.} $S=(26,22,22,18)$ for $H=64$, giving $C/C_{\mathrm{full}}=0.120$. Each layer truncates the DCT spectrum to $s_l\times s_l$ and applies the reduced inverse basis as in Resolution Layers~\cite{harper2026}.

\textbf{Soft.} $S$ is learned with the sigmoid mask on the full $H\times H$ tensor as in DNArch~\cite{dnarch}. The budget constrains the retained spectrum but not feature-map size; this requires about $6.5\times$ the forward FLOPs and $2.5$--$3.6\times$ the training time per step of fixed.

\textbf{Crop.} We extend~\cite{dnarch} by keeping only the first
$K_l=\lceil s_l+\tau\ln((1-\epsilon)/\epsilon)\rceil$ coefficients, with $\epsilon=0.01$, and using a $K_l$-point inverse basis. Discarded coefficients have mask values below 0.01, leaving gradients to $s_l$ essentially unaffected while shrinking the feature maps. The retained area $\sum_l K_l^2/(LH^2)$ averages about 0.24 for both kernel types, roughly twice the budget value because the mask tail is retained; the kernel types therefore have approximately matched retained area. Sec.~4.1 also compares crop with an equal-area fixed schedule. We additionally test DiffStride's linear ramp~\cite{diffstride} of width 4.

In all learnable variants each $s_l$ is clipped to $[2, H]$ and initialized at $1.3\times$ the fixed schedule.

\subsection{Spectral Kernels}

The spatial kernel is a standard $3 \times 3$ convolution. The spectral kernel, SpecMLP, generates the kernel directly in the frequency domain, similar to~\cite{harper2025scaling}: each normalized frequency coordinate $(u, v)$ is passed through a two-layer MLP (hidden width 64) that outputs $C_{\mathrm{in}} \times C_{\mathrm{out}} \times 2$ values, the real and imaginary parts of the kernel at that frequency, which multiply the real FFT of the input before an inverse transform. SpecMLP is thus the frequency-domain analogue of CKConv~\cite{ckconv}.

The backbone has a stem convolution followed by four stages, and the kernel under study replaces the convolution in the first stage. We use $F = 16$ channels for the spatial kernel and $F = 8$ for SpecMLP ($20{,}974$ and $24{,}790$ parameters); Sec.~4.3 controls for width and parameter count, and Sec.~4.4 scales both models.

\section{Experimental Results}
\label{sec:exp}

We use EuroSAT RGB~\cite{helber2019eurosat}, CIFAR-10~\cite{krizhevsky2009learning}, and STL-10~\cite{coates2011analysis}. For EuroSAT, the public 60/20/20 split is used, with its 16,200-image training partition further split 80/20 into 12,960 training and 3,240 validation images; the 5,400 test images remain held out. For CIFAR-10, 12,960 class-balanced images from the official training split are used for training, while the official test split is partitioned into 3,240 validation and 6,760 held-out test images. The 13,000 labeled STL-10 images are split class-balanced into 10,000/1,500/1,500 training/validation/test images.

Models are trained for 30 epochs with Adam unless otherwise noted
(batch size 32, initial learning rate $5\times10^{-4}$ with cosine
decay; $100\times$ larger for $S$). Each configuration uses 10 seeds unless noted. Table~1 and Fig.~1 use held-out test accuracy; secondary analyses use validation when test outputs were unavailable, with a common split used within each grouped comparison. Validation results are marked. We report $\mathrm{Gain}(k)=A_{\mathrm{learn}}^{(k)}-A_{\mathrm{fixed}}^{(k)}$ and $\Delta_{\mathrm{gain}}=\mathrm{Gain}(\mathrm{spectral})-\mathrm{Gain}(\mathrm{spatial})$. Paired $t$-tests are used across seeds; unpaired Welch tests agree on all main results. The four main crop comparisons remain significant after Holm correction; secondary tests are uncorrected. Absolute accuracy is limited by the small backbones and short training; the largest EuroSAT model reaches 0.95 (Sec.~4.4).

\subsection{Main Result: Learnable Downsampling Gains}

\begin{table}[t]
\centering
\caption{Gain from making $S$ learnable at $p^\star=0.12$, held-out test accuracy ($n=10$). The first two columns are the gains under crop; the last three are $\Delta_{\mathrm{gain}}$ under each implementation. EuroSAT-S uses the smaller models ($2.1\times10^4$ spatial and $2.5\times10^4$ spectral parameters); all other rows use approximately $5\times10^4$ parameters. $^\ast$: uncorrected $p<0.05$; all four crop comparisons remain significant after Holm correction.}
\label{tab:main}
\setlength{\tabcolsep}{2.5pt}
\begin{tabular}{l|cc|ccc}
\hline
 & \multicolumn{2}{c}{Gain (\emph{crop})} & \multicolumn{3}{c}{$\Delta_{\mathrm{gain}}$} \\
 & spatial & spectral & \emph{crop} & ramp & \emph{soft} \\
\hline
EuroSAT-S & $+0.009$ & $+0.022$ & $+0.013^*$ & $+0.006$   & $+0.020^*$ \\
EuroSAT   & $+0.004$ & $+0.021$ & $+0.016^*$ & $+0.012^*$ & $+0.022^*$ \\
CIFAR-10  & $-0.007$ & $+0.011$ & $+0.018^*$ & $+0.002$   & $+0.038^*$ \\
STL-10    & $+0.003$ & $+0.031$ & $+0.028^*$ & $+0.008$   & $+0.049^*$ \\
\hline
\end{tabular}
\end{table}

We first compare the accuracy gains for downsampling with spatial and spectral kernels. Table~1 and Fig.~1 report the accuracy gain from learnable $S$ under crop resizing. In all four settings, the spectral kernel gains more than the spatial one: the $\Delta_{\mathrm{gain}}$ is +0.013 to +0.028 on the test sets ($p \leq 0.02$) and +0.020 to +0.025 on validation. The spatial kernel gains little ($-0.007$ to +0.009), whereas the spectral kernel gains +0.011 to +0.031.

Because crop retains about 0.24 of the feature-map area, roughly twice the 0.12 of the fixed schedule, we also compare against a fixed schedule with matched retained area ($p^\star = 0.25$). Increasing the fixed schedule to this area without learning gives a $\Delta_{\mathrm{gain}}$ of $+0.007$ to $+0.018$ on validation, significant on EuroSAT and STL-10, showing that part of its gain under crop comes from retaining more spatial area. However, relative to this matched-area baseline, learned allocation still produces a larger gain for the spectral kernel in every setting ($\Delta_{\mathrm{gain}} = +0.006$ to +0.015), significant in two of four cases (Table~2).

\subsection{Effect of the Resizing Implementation}

The $\Delta_{\mathrm{gain}}$ depends on how resizing is implemented (Table~\ref{tab:main}). \emph{Soft} gives the largest values ($+0.020$ to $+0.049$) at about $6.5\times$ the FLOPs of \emph{fixed}, and on CIFAR-10 and STL-10 much of this comes from it lowering the spatial kernel ($-0.050$, $-0.037$) rather than from spectral gains. The tighter ramp crop ($s_l + 4$) yields smaller $\Delta_{\mathrm{gain}}$, significant only for EuroSAT with the larger models. A CIFAR-10 threshold sweep gives $\Delta_{\mathrm{gain}}=+0.012,+0.018,+0.019$ for $\epsilon=0.1,0.01,0.001$, respectively, suggesting saturation once sufficient spectrum above the cutoff is retained. We use $\epsilon=0.01$ for the main crop comparison.

\subsection{Other Factors}

\begin{table}[t]
\centering
\caption{Controls at $p^\star = 0.12$. Top: crop compared with a fixed schedule of matched retained area ($p^\star = 0.25$), validation. Middle: additional controls, test unless marked ``val.''. Bottom: a fixed high-ratio schedule ($s_1/s_2 = 2.06$) against the default fixed schedule, test.}
\label{tab:controls}
\setlength{\tabcolsep}{4pt}
\begin{tabular}{lc}
\hline
Control & $\Delta_{\mathrm{gain}}$ ($p$) \\
\hline
Equal area, EuroSAT-S                        & $+0.006$ ($0.08$) \\
Equal area, EuroSAT                          & $+0.013$ ($0.004$) \\
Equal area, CIFAR-10                         & $+0.015$ ($0.03$) \\
Equal area, STL-10                           & $+0.009$ ($0.13$) \\
\hline
Spatial baseline, $F = 8$ ($12.7$k params)   & $+0.020$ ($0.002$) \\
Spatial baseline, $F = 19$ ($25.3$k params)  & $+0.023$ ($0.002$) \\
CIFAR-10 at native $32 \times 32$ (val.)     & $+0.031$ ($0.001$) \\
\hline
Fixed high-ratio schedule ($n = 30$)         & $+0.014$ ($0.0003$) \\
\hline
\end{tabular}
\end{table}

The remaining rows of Table~\ref{tab:controls} rule out simpler explanations. Narrowing the spatial baseline or matching its parameter count leaves $\Delta_{\mathrm{gain}}$ unchanged. Training CIFAR-10 at its native resolution, which avoids the empty spectrum introduced by upsampling, gives the same result as the upsampled version. Training CIFAR-10 for 100 epochs raises fixed-schedule test accuracy by about $0.06$ and retains a significant test-set $\Delta_{\mathrm{gain}}$ of +0.014 ($n=10$, $p=0.0002$), versus +0.018 at 30 epochs. A CKConv-style kernel generated by a SIREN~\cite{sitzmann2020siren} over spatial offsets gains $+0.014$ from learnable $S$, between the $3 \times 3$ kernel ($+0.007$) and SpecMLP ($+0.027$); the ordering is significant on validation but not on test. Finally, imposing without learning a fixed schedule with the high first-stage ratio that the spectral filters learn ($s_1/s_2 = 2.06$, at the area of fixed) improves the spectral kernel by +0.018 and the spatial kernel by +0.004 ($n = 30$), a $\Delta_{\mathrm{gain}}$ of +0.014 ($p = 0.0003$), similar in size to that of crop for these models (+0.013). On EuroSAT, imposing the allocation that the spectral filters learn thus reproduces a $\Delta_{\mathrm{gain}}$ of similar size without learning.

\subsection{Budget and Model Size}

\begin{figure}[t]
\centering
\includegraphics[width=\linewidth]{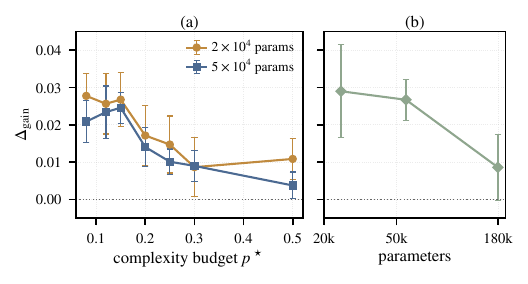}
\caption{$\Delta_{\mathrm{gain}}$ on EuroSAT under \emph{soft}. (a) Across seven budgets at two model sizes ($n = 20$ per point, validation). (b) Across model size at $p^\star = 0.12$ ($n = 10$, validation). Error bars are 95\% confidence intervals.}
\label{fig:trends}
\end{figure}

Fig.~\ref{fig:trends} varies budget and model size under \emph{soft}. The $\Delta_{\mathrm{gain}}$ is largest under tight budgets (about $+0.026$ for $p^\star \le 0.15$) and falls to $+0.011$ and $+0.004$ at $p^\star = 0.50$; all 14 points are significant. Scaling from the smaller models ($\approx 2$--$2.5\times10^4$ parameters) to $1.8\times10^5$ parameters, the validation $\Delta_{\mathrm{gain}}$ persists but decreases (+0.029, +0.027, +0.009); the test-set differences are significant (+0.020, +0.022, +0.008; $p=0.013$, $p<10^{-4}$, and $p=0.014$, respectively). At the largest size, spectral and spatial accuracies reach 0.951 and 0.921. The scale trend also holds under crop on validation (+0.024, +0.021, +0.007; all significant), but the budget trend does not: under crop the $\Delta_{\mathrm{gain}}$ is not significant at $p^\star = 0.25$ or 0.50. Thus, $\Delta_{\mathrm{gain}}$ persists across model sizes but weakens with capacity and depends on the resizing implementation.

\subsection{Learned Allocation and Kernel Size}

\begin{figure}[t]
\centering
\includegraphics[width=\linewidth]{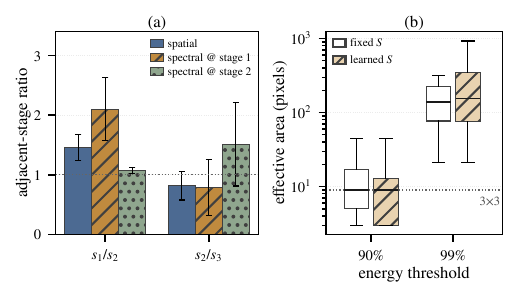}
\caption{Learned allocation and kernel size on EuroSAT at $p^\star = 0.12$. (a) Mean adjacent-stage resolution ratios under \emph{soft} over 10 seeds (separate runs), with the spectral kernel at stage 1 or 2; error bars are one standard deviation. (b) Effective area of the SpecMLP filters (640 per configuration) on the layer's own grid at two energy thresholds; boxes show quartiles, whiskers the 5th--95th percentiles.}
\label{fig:alloc}
\end{figure}

On EuroSAT, the spectral filters allocate relatively more resolution to the stage in which they operate: under \emph{soft} they learn $s_1/s_2 = 2.09 \pm 0.54$ against $1.34 \pm 0.11$ for spatial filters ($p = 4 \times 10^{-4}$), and moving the spectral kernel to the second stage shifts the allocation preference accordingly (Fig.~\ref{fig:alloc}(a)). The same pattern appears under \emph{crop} on EuroSAT ($K_1/K_2$ of $1.8$ against $1.2$) but only weakly on the other datasets, so we report it as a property of the EuroSAT solutions, not a mechanism. The learned filters are compact: the median filter places 83\% of its energy within $3 \times 3$, and its effective area does not differ between \emph{fixed} and \emph{soft} ($p = 0.84$; Fig.~\ref{fig:alloc}(b)). With the larger EuroSAT models, the qualitative pattern is robust to initialization: under \emph{crop}, initializing $S$ at $1.0\times$, $1.3\times$, or $1.6\times$ the fixed schedule yields test-set $\Delta_{\mathrm{gain}}=+0.021,+0.016,+0.015$, respectively, while preserving a larger learned first-to-second-stage cutoff ratio for spectral filters.

\section{Conclusion}

We studied how learnable resizing interacts with convolutional filter parameterization. Under matched cutoff-frequency budgets, spectral and $3\times3$ spatial filters respond differently to learned resolution allocation, with larger gains for spectral filters across three datasets and two model sizes. The effect weakens with model size and depends on the resizing implementation.  The effect is modest and largest for small models under tight budgets.

The models remain relatively small despite the scale study, and we test one backbone with a single spectral parameterization in a single layer. Crop retains about twice the area of the fixed schedule, explaining part of the observed gain. The remaining $\Delta_{\mathrm{gain}}$ is not fully explained; measuring how spectral filters use frequencies above the cutoff is a natural next step.

\section*{Acknowledgments}
Computations used SMU's ManeFrame III. Claude (Anthropic) assisted with drafting/editing Secs.~2--5, figure/table captions, and experiment/plotting scripts; the authors designed and ran the study and verified all text, code, and reported results.

\bibliographystyle{IEEEbib}
\bibliography{strings,refs}

\end{document}